\documentclass{article}
\usepackage{spconf,amsmath,graphicx}
\usepackage{multirow}
\usepackage{multicol}

\usepackage{xcolor} 
\usepackage{hyperref}

\title{Real-Time Wheel Detection and Rim Classification\\in Automotive Production}

\name{Roman Staněk~$^{1}$,\thanks{This work was partially supported by the Czech Science Foundation, grant no. GA21-03921S, and by the \textit{Praemium Academiae} awarded by the Czech Academy of Sciences.}
Tomáš Kerepecký~$^{2,3}$,
Adam Novozámský~$^{3}$,
Filip Šroubek~$^{3}$,
Barbara Zitová~$^{3}$,
Jan Flusser~$^{3}$}
\address{
\normalsize {$^{1}$Charles University, Czechia, $^{2}$Czech Technical University in Prague, Czechia}\\
\normalsize{$^{3}$Institute of Information Theory and Automation, The Czech Academy of Sciences, Czechia}
}

\begin{document}
\maketitle
\begin{abstract}
This paper proposes a novel approach to real-time automatic rim detection, classification, and inspection by combining traditional computer vision and deep learning techniques. At the end of every automotive assembly line, a quality control process is carried out to identify any potential defects in the produced cars. Common yet hazardous defects are related, for example, to incorrectly mounted rims. Routine inspections are mostly conducted by human workers that are negatively affected by factors such as fatigue or distraction. We have designed a new prototype to validate whether all four wheels on a single car match in size and type. 
Additionally, we present three comprehensive open-source databases, CWD1500, WHEEL22, and RB600, for wheel, rim, and bolt detection, as well as rim classification, which are free-to-use for scientific purposes.
\end{abstract}
\vspace{-0.2cm}
\section{Introduction}
\vspace{-0.2cm}
\label{sec:introduction}
In 2021, global motor vehicle production was estimated to be over 80 million vehicles~\cite{2023oica}. Most quality check tasks are performed by trained workers, who can be affected by many negative factors, which reduces the reliability of the inspection. This tedious work provides a significant opportunity for automation through computer vision, which has the potential to lower the cost of the overall process and, at the same time, achieve superior accuracy. Despite the prevalence of automated computer vision tasks, the quality control of rim mounting inaccuracies is still done manually. This paper aims to design a real-time system to ensure that all four rims on a car are of the same size and type, which is a crucial factor for maintaining car stability and passenger safety.

There are limited studies focused on wheel detection and, to the best of our knowledge, a lack of literature regarding rim classification. In this regard, we present a novel approach that constitutes the first comprehensive pipeline for joint wheel detection, classification, and size estimation. \vspace{10pt}

\begin{figure}[t]
  \renewcommand{\arraystretch}{0.4}
  \begin{center}
  \begin{tabular}{@{}c@{}c@{}}
    \includegraphics[width=0.35\linewidth]{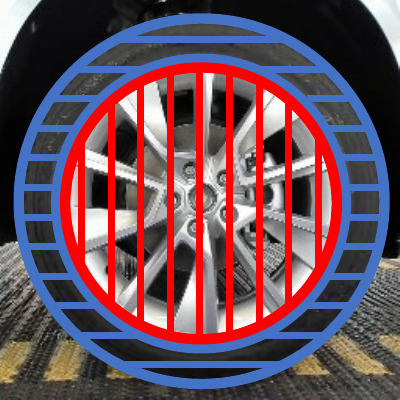}\hspace{2px} &
    \includegraphics[width=0.35\linewidth]{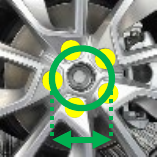}\hspace{2px}
  \end{tabular}
  \vspace{-0.3cm}
  \caption{Visualization of the main wheel parts. The tyre is marked in blue, the rim in red, the wheel bolt centers in yellow, and the diameter of the pitch circle in green.}
  \label{fig:wheel}
\vspace{-0.7cm}
\end{center}
\end{figure}

\noindent \textbf{Contributions.}
\textbf{(1)}~we propose a novel real-time rim detection, classification, and inspection approach and design a prototype ready-to-use in automotive production; \textbf{(2)}~we contribute three new  publicly available datasets, namely \textbf{CWD1500} for car and wheel detection, \textbf{WHEEL22} for rim classification and \textbf{RB600} for bolt detection.

\vspace{-0.3cm}
\section{Related Work}
\vspace{-0.2cm}
\label{sec:related_work}
We present an overview only in the area of wheel detection, as there is no public research on rim classification. These studies predominantly employ the \textit{Hough transform~(HT)}~\cite{2015hassanein} and use machine learning to determine the presence of a wheel. 

A comprehensive introduction to car and wheel detection is in ~\cite{2015hassanein}, which presents a three-stage approach for detecting car contours from side view and identifying wheels using HT and \textit{SURF descriptors}~\cite{2006bay}. They use heuristics and a \textit{Snake algorithm}~\cite{1988kass} to improve results, but the results are inconclusive due to a small dataset (100 images) and white background.

The Master thesis~\cite{2013hultstrom} detects wheels utilizing real-world recordings and using \textit{Local binary patterns}~\cite{1996ojala} and \textit{Random forest classifier}~\cite{1995ho}. Another work~\cite{2011chavez} identifies 14 regions of interest in vehicles from a side view, including wheels, using a classifier trained on Haar-like features~\cite{2001viola} and HT.

The paper~\cite{2015antonov} uses \textit{Fast~HT}~\cite{1986li} to detect wheels in a deployed industrial vehicle classification system in Russia, filtering the Hough space to avoid false positives.

\vspace{-0.2cm}
\section{Method}
\vspace{-0.2cm}
\label{sec:method}
Our method consists of several building blocks in which standard computer vision methods are complemented with deep learning methods. We start with the creation of three datasets and then describe the car and wheel detection, rim classification, and finally, rim-size estimation. The terms \textit{wheel}, \textit{rim}, and \textit{tyre} can be ambiguous in the common language. In this paper, we refer to a wheel as the combination of rim and tyre. A rim is a rigid core of the wheel usually made from metal. A tyre is mounted on the rim and ensures good contact with the surface under the car. It is usually made of rubber-like material. For illustration, see Figure~\ref{fig:wheel}. The primary purpose of the rim is to provide rigid support for the tyre and to transmit forces that affect the movement of the vehicle, such as the rotational force from the engine to the tyre.~\cite{2018Leister}.

\begin{figure}[!bp]\centering
\vspace{-0.1cm}
\includegraphics[width=0.95\linewidth]{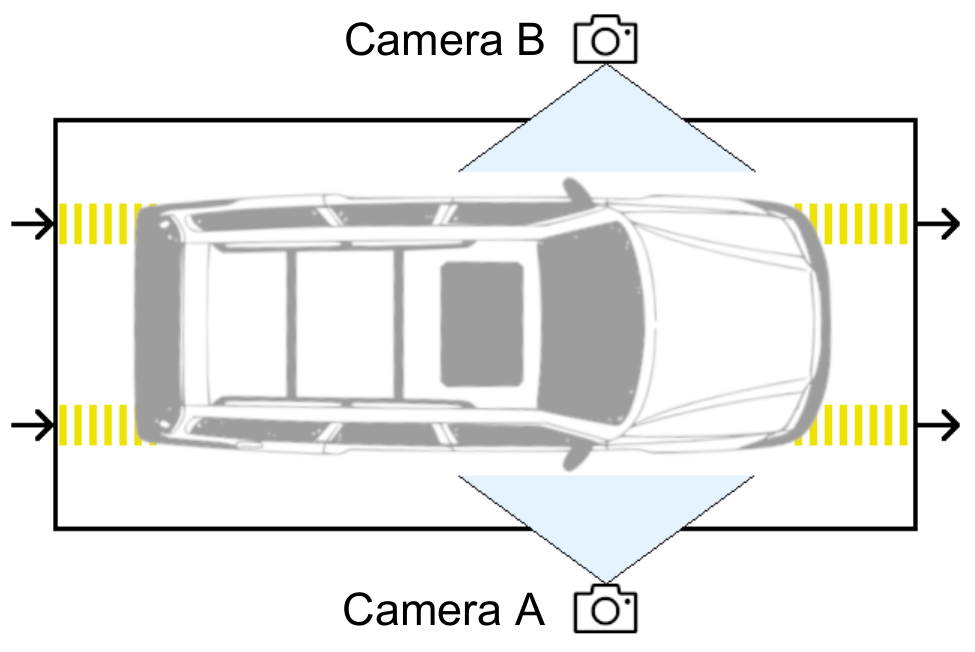}
\vspace{-0.5cm}
\caption{Top-Down Camera Setup: The cameras were placed on either side of the conveyor belt, with Camera A capturing cars from the right and Camera B capturing them from the left.}
\label{fig:cameras}
\vspace{-0.1cm}
\end{figure}

\begin{figure}[h]\centering
\includegraphics[width=0.95\linewidth]{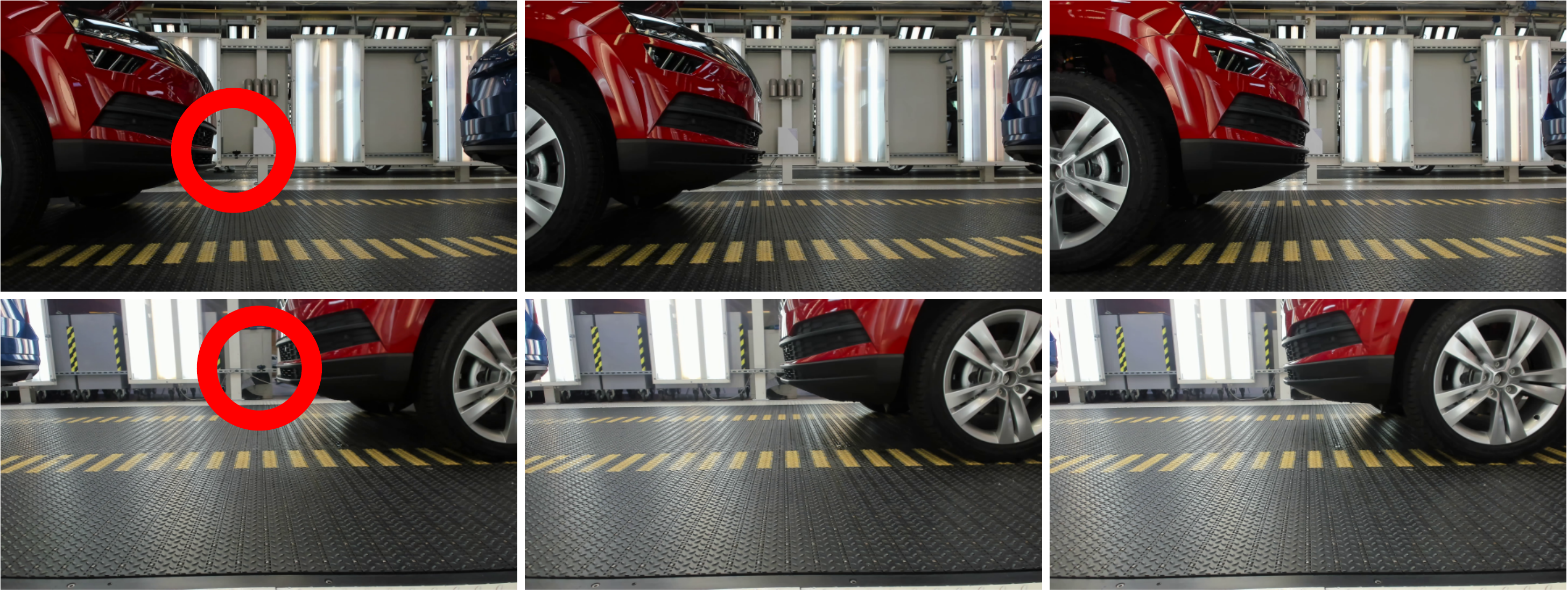}
\vspace{-0.3cm}
\caption{Sequences of 3 subsequent frames from Camera A (top) and Camera B (bottom) illustrate the conveyor belt movement speed - approximately 11.5 cm/s. The red circles in the left column show the opposite camera.}
\label{fig:camview}
\vspace{-0.5cm}
\end{figure}

\vspace{-0.3cm}
\subsection{Datasets}
\label{subsec:datasets}
\begin{figure}[!t]
\vspace{0.3cm}
  \renewcommand{\arraystretch}{0.4}
  \begin{center}
  \begin{tabular}{@{}c@{}c@{}}
    \includegraphics[width=0.45\linewidth]{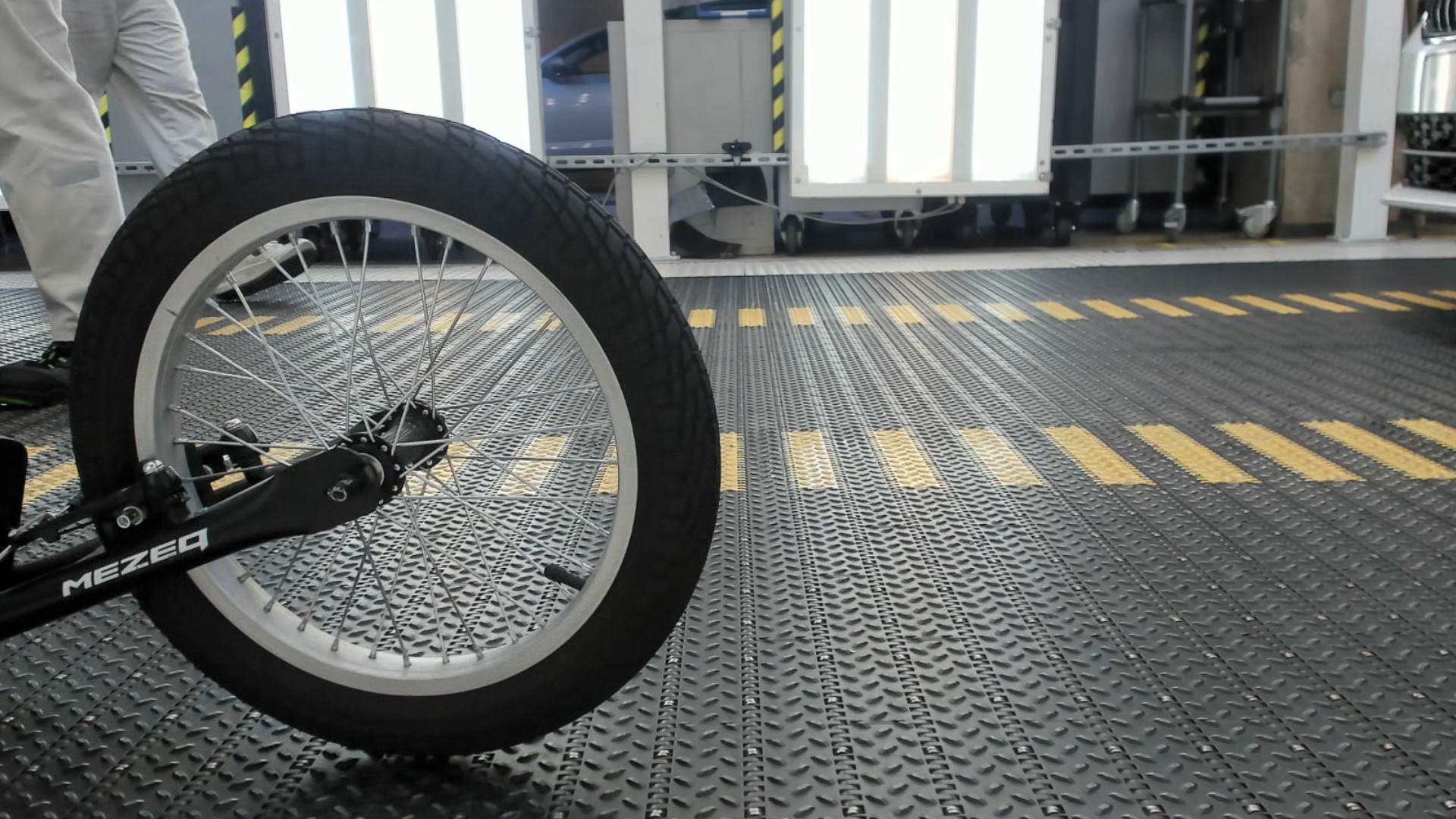}\hspace{2px} &
    \includegraphics[width=0.45\linewidth]{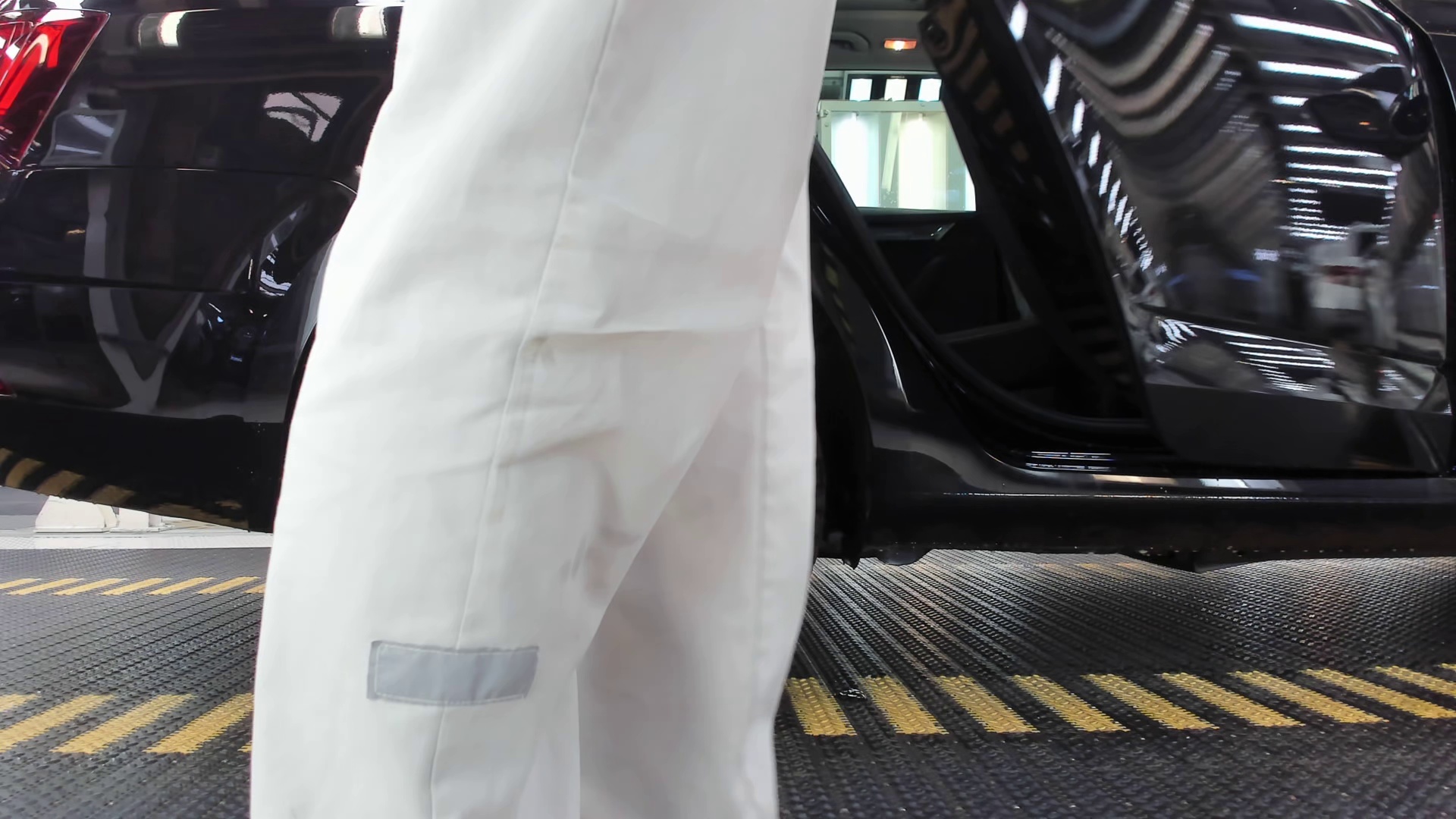}\hspace{2px}
  \end{tabular}
  \vspace{-0.3cm}
  \caption{Examples of problematic objects in the scene: a scooter wheel is not the object of interest, and a staff standing in front of the car completely occluding the wheel.}
  \label{fig:objects}
    \vspace{-0.5cm}
\end{center}
\end{figure}

\vspace{-0.2cm}
All data were collected at the Škoda Auto factory in Mladá Boleslav, where quality control is carried out. This company permitted us to install the monitoring equipment to gather car data and use them for research purposes. Cars are stationary on a slowly moving conveyor belt that is well-lit by multiple sources. The cameras were arranged on both sides of the conveyor belt; Camera A recorded cars from the right side, and Camera B from the left. A top-down schema of the setup is shown in Figure~\ref{fig:cameras}. Images of the scene taken by individual cameras simultaneously are shown in Figure~\ref{fig:camview}. We employed standard Logitech BRIO 4K Stream Edition cameras with the same configuration as in~\cite{2020novozamsky}. The data collection script ran continuously for several days at a rate of one frame per second in FullHD quality, recording data in ten-minute bursts. Every frame was captured as Motion JPEG, and the whole time-lapse video was encoded in HEVC H.265. The data was gathered during standard work shifts; people and objects can move in the scene; see Figure~\ref{fig:objects}. Thirty-four hours of collected data from both cameras were obtained after removing unusable video segments.
It is necessary to provide training data with accurately labeled objects to train a neural network for object detection or tuning parameters for standard algorithms such as HT. We manually annotated the cars and wheels by drawing bounding boxes in the training images using the \textit{Computer Vision Annotation Tool(CVAT)}~\cite{2022cvat}. The first dataset, \textbf{CWD1500}, is designed for car and wheel detection and includes 1000 training frames, 250 validation frames, and 250 testing frames. The training set also includes 91 unlabelled images to prevent false positives during training. 

\begin{figure*}[h]
  \renewcommand{\arraystretch}{0.4}
  \begin{center}
  \begin{tabular}{@{}c@{}c@{}c@{}c@{}c@{}c@{}c@{}c@{}c@{}c@{}c@{}}    
    \small C00 &\small C01 &\small C02 &\small C03 &\small C04 &\small C05 &\small C06 &\small C07 &\small C08 &\small C09 &\small C10\\
    \includegraphics[width=0.09\textwidth]{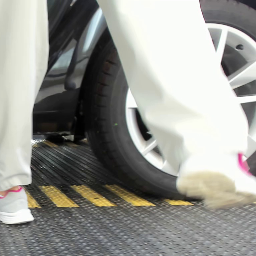}&
    \includegraphics[width=0.09\textwidth]{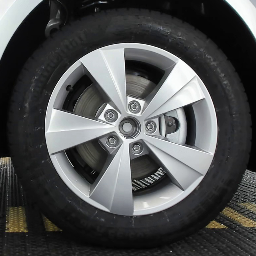}&    
    \includegraphics[width=0.09\textwidth]{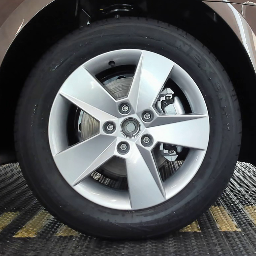}&
    \includegraphics[width=0.09\textwidth]{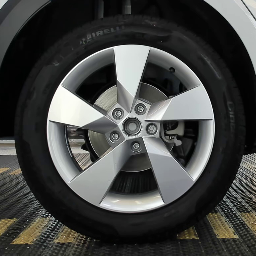}&
    \includegraphics[width=0.09\textwidth]{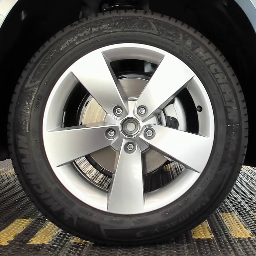}&
    \includegraphics[width=0.09\textwidth]{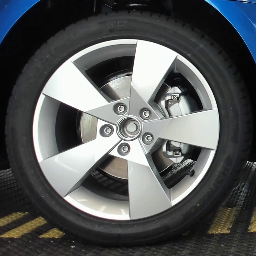}&
    \includegraphics[width=0.09\textwidth]{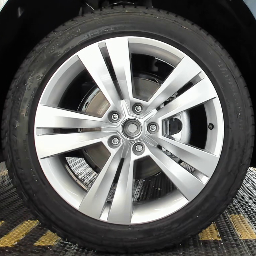}&
    \includegraphics[width=0.09\textwidth]{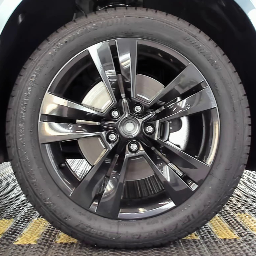}&
    \includegraphics[width=0.09\textwidth]{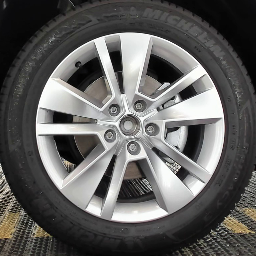}&
    \includegraphics[width=0.09\textwidth]{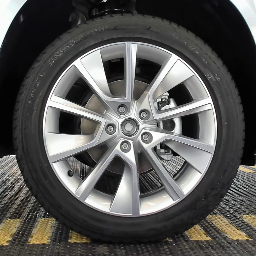}&
    \includegraphics[width=0.09\textwidth]{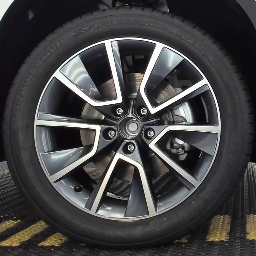}\\  
    \small C11 &\small C12 &\small C13 &\small C14 &\small C15 &\small C16 &\small C17 &\small C18 &\small C19 &\small C20 &\small C21\\
    \includegraphics[width=0.09\textwidth]{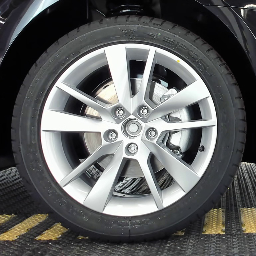}&
    \includegraphics[width=0.09\textwidth]{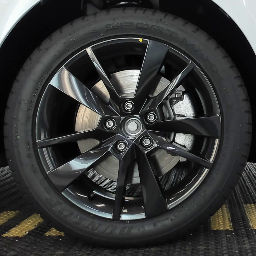}&
    \includegraphics[width=0.09\textwidth]{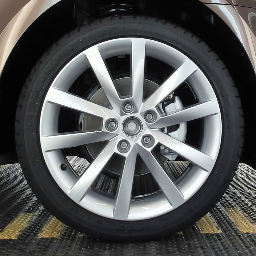}&
    \includegraphics[width=0.09\textwidth]{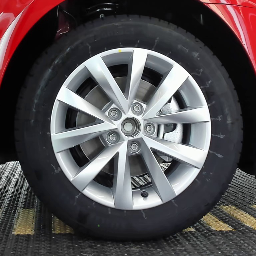}&
    \includegraphics[width=0.09\textwidth]{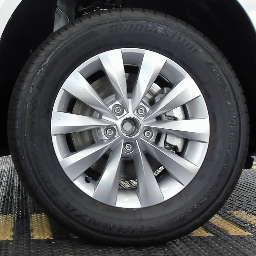}&
    \includegraphics[width=0.09\textwidth]{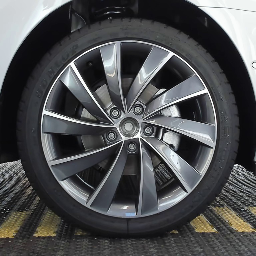}&
    \includegraphics[width=0.09\textwidth]{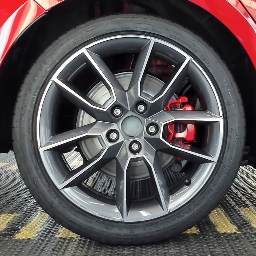}&
    \includegraphics[width=0.09\textwidth]{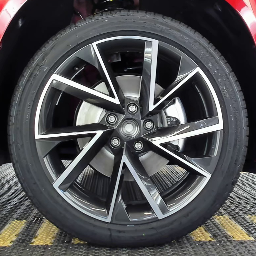}&
    \includegraphics[width=0.09\textwidth]{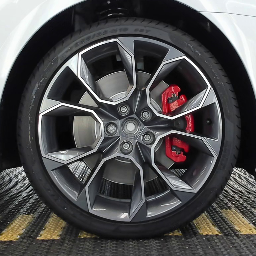}&
    \includegraphics[width=0.09\textwidth]{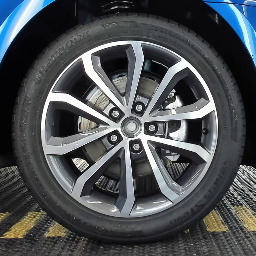}&
    \includegraphics[width=0.09\textwidth]{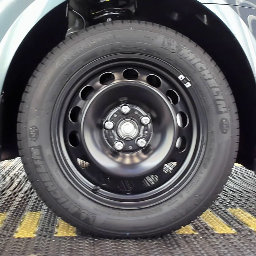}\\
  \end{tabular}
  \vspace{-0.3cm}
  \caption{WHEEL22 dataset: 1+21 rim categories. C00 category is used for handling occlusions.}
  \label{fig:classes}
    \vspace{-0.5cm}
\end{center}
\end{figure*}

We employed the YOLO~\cite{2015redmon} detection network, which had been trained using the CWD1500 dataset, to identify the location of all rims in the collected videos. Subsequently, each frame was processed by cropping it to a square shape centered around the detected bounding box and resized to 256 x 256 pixels. It is important to note that some of the identified rims were not entirely captured in the pictures, which assisted in generalizing the learning process of classification, since it functioned as cropping in standard data augmentation. We manually identified 31 classes of rims based on differences in shape and color of detected wheels. For ten classes the number of representative samples was too low (less than 300). Therefore, we used only 21 classes for further analysis. We randomly selected 100 training samples for each class, 25 validation samples, and 25 test samples. It should be noted that images of one car will only appear in one set (training, validation, or test). We added one special class to include cases when the detector returns a candidate that cannot be classified even by a human. One example per class is shown in Figure~\ref{fig:classes}. We refer to this comprehensive labeled dataset containing 3300 rim images as \textbf{WHEEL22}.

The last dataset was created for detecting five wheel bolts. It contains 400 images for training, 100 for testing, and 100 for validation. The rims and bolts were also manually annotated using CVAT. This dataset is called \textbf{RB600}.

\begin{figure*}[h]
\begin{center}
  \includegraphics[width=0.95\textwidth]{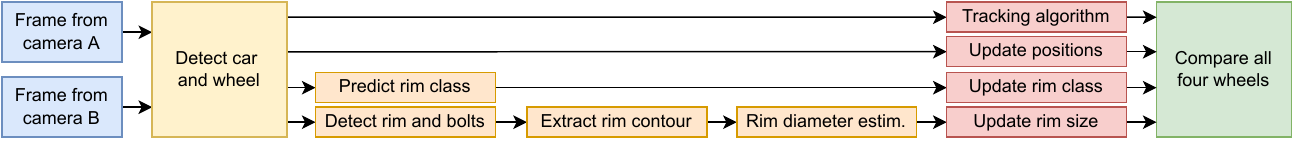}
  \vspace{-0.3cm}
  \caption{Data flow in the proposed prototype.}
\label{fig:flowchart}
\vspace{-0.5cm}
\end{center}
\end{figure*} 

\vspace{-0.3cm}
\subsection{Car and wheel detection}
\label{subsec:detection1}
\vspace{-0.2cm}
This section is split into two subsections. The first part deals with research utilizing traditional computer vision techniques, while the latter focuses on convolutional neural networks. A comparison of the two approaches is presented in Table~\ref{table:detection}. There are \textit{precision}~(P), \textit{recall}~(R), and \textit{mean average precision} (mAP) to evaluate object detection performance. All of them are calculated using the \textit{Intersection-over-Union} (IoU) threshold of 50\%. If a range is specified, such as mAP@.5:.95, it indicates the average mAP over an IoU range from 0.5 to 0.95 with step size 0.05. From the results, it is evident that the deep learning method outperforms traditional methods.

\textit{\textbf{a)Traditional Computer Vision:}} We chose HT  as the most appropriate method based on the related work presented in Section~\ref{sec:related_work}. HT is a widely accepted technique for detecting geometric shapes that a limited number of parameters can describe. We utilized an implementation provided by~\cite{1990hough} that was optimized for minimizing memory usage. To enhance its performance, we carried out several preprocessing steps. The image was first downscaled, converted to grayscale, and then blurred using a Gaussian filter to reduce noise and eliminate high-frequency components.
The results obtained by measuring performance on the validation dataset of CWD1500 are summarized in Table~\ref{table:detection} (first line) and are satisfactory for a baseline method. The true positives do not always match the rim contours, as the car wheels are not always perpendicular to the cameras' optical axes.
The false negatives in the detection can be attributed to three reasons: dim rims, partially occluded rims, and rims partially out of the camera field of view. In the third group, the wheels are missing more than half of their area, which is hard for HT to detect.

\textit{\textbf{b)Deep Learning Approach:}} Standard deep-learning detection methods include approaches such as U-net~\cite{2015ronneberger}, Mask-RCNN~\cite{2017he}, and YOLO~\cite{2015redmon}. We chose 
YOLOv5s specification~\cite{2022jocher} with 7M parameters and pre-trained on the COCO val2017 dataset. The model was re-trained for 50 epochs using a 2:1:1 split of the CWD1500 dataset. The estimated inference time on a single frame with YOLOv5s is around 26 ms, which is still sufficiently fast  for our purposes (HT takes only 8ms).

\vspace{-0.4cm}
\subsection{Rim classification}
\label{subsec:classification}
\vspace{-0.2cm}
We compared traditional techniques with deep learning approaches as in the previous section on wheel detection.

\textit{\textbf{a)Traditional Computer Vision:}} As the traditional computer vision technique, we combined a \textit{Histogram of Oriented Gradients}~(HOG)~\cite{2005dalal} with a \textit{Support Vector Machines}~(SVM)~\cite{1995cortes} classifier. HOG is a feature descriptor used in computer vision for object detection. It represents the orientation of intensity gradients in an image, dividing the image into small cells and counting the gradient directions in each cell. The results of this analysis are then compiled into a histogram, which serves as a descriptor of the object's appearance.
The 'orientation' parameter defines how many bins the histogram has in each cell. The parameter 'pixels per cell' describes the size of one cell in pixels. The results for three algorithm settings on WHEEL22 dataset are summarized in Table~\ref{table:hog}. In this case, the maximum achieved accuracy below $0.75$ is insufficient. 

\begin{table}[b]
\vspace{-0.2cm}
\begin{center}
\resizebox{\columnwidth}{!}{%
\begin{tabular}{llcccccc}
\multicolumn{2}{}{} \textbf{Method} & \textbf{Class} & \textbf{Instances}& \textbf{P} & \textbf{R} & \textbf{mAP@.5} & \textbf{mAP@.5:.95} \\
\hline\hline
\multicolumn{2}{}{} HT  & wheel & 182 & 1.000 & 0.703 & --- & --- \\
\hline
    \multirow{4}{*}{\rotatebox[origin=c]{90}{YOLO}}  & \multirow{2}{*}{Sec.~\ref{subsec:detection1}} & wheel & 182 & 0.983 & 0.970 & 0.993 & 0.962 \\
     & & car & 300 & 0.983 & 0.980 & 0.993 & 0.928 \\   
\cline{2-8}
    & \multirow{2}{*}{Sec.~\ref{subsec:detection2}} & bolt & 475 & 1.000 & 0.998 & 0.995 & 0.651 \\
    & & rim & 95 & 0.984 & 1.000 & 0.995 & 0.993 \\     
\end{tabular}
}
\vspace{-0.3cm}
\caption{The performance of the detection method was evaluated on test data, comparing two models: HT and YOLOv5s, using the CWD1500 dataset. Additionally, the bottom section of the results presents the outcomes of the same YOLO architecture trained and tested on the RB600 dataset.}
\label{table:detection}
  \vspace{-0.3cm}
\end{center}
\end{table}

\begin{table}[h]
\begin{center}
\resizebox{\columnwidth}{!}{%
\begin{tabular}{cccc}
\textbf{Orientation} & \textbf{Pixels per cell} & \textbf{Accuracy}& \textbf{Features} \\
\hline
    9 & 8x8 & 0.643 & 73K \\    
    13 & 24x24 & \textbf{0.744 }& 7.5K \\
    16 & 24x24 & 0.718 & 9.2K 
\end{tabular}
}
\vspace{-0.3cm}
\caption{Results of HOG with multiple variations of parameters. The column \textit{Features} describes the number of features generated per image.}
\label{table:hog}
  \vspace{-0.5cm}
\end{center}
\end{table}

\begin{table}[h]
\vspace{-0.1cm}
\begin{center}
\resizebox{\columnwidth}{!}{%
\begin{tabular}{cccc}
\textbf{Unfrozen layers} & \textbf{Trainable param.} & \textbf{Accuracy}& \textbf{Train time [s]} \\
\hline
    2 & 2.5K & 0.956 & 574 \\    
    10 & 893.2K & 0.989 & 659 \\
    25 & 1.5M & \textbf{0.995} & 1103 \\
    237 & 4M & 0.989 & 1848
\end{tabular}
}
\vspace{-0.3cm}
\caption{Transfer learning results using EfficientNet with various unfrozen layer configurations.}
\label{table:cnn}
  \vspace{-0.5cm}
\end{center}
\end{table}

\textit{\textbf{b)Deep Learning Approach:}} Here we will concentrate on transfer learning and EfficientNet~\cite{2019tan}, a high-performing classification network. Transfer learning is a technique where a model pre-trained on a large dataset, such as ImageNet~\cite{2015russakovsky}, is then used for training on a smaller dataset. This process accelerates and improves training by leveraging the learned representations from the pre-trained model. The layers of the pre-trained model are divided into two categories: frozen and trainable. The frozen layers remain unchanged, while the trainable layers are modified during the training. The number of trainable layers varies depending on the model, dataset size, and complexity. Individual runs with the number of unfrozen layers and achieved validation accuracy are in Table~\ref{table:cnn}. The findings show that it is sufficient to train 25 layers, for which we reach a close-to-optimal model while training time is still moderate. The inference time of 60 ms per image is also favorable for our intended application. The confusion matrix for the test data shows similarity to the identity matrix, except for three classes. Specifically, C08 had two misrecognized representatives, C09 had four, and C14 had one, which resulted in an overall accuracy of 98.72\%.

\begin{figure}[!bp]\centering
\vspace{-0.4cm}
\includegraphics[width=0.95\linewidth]{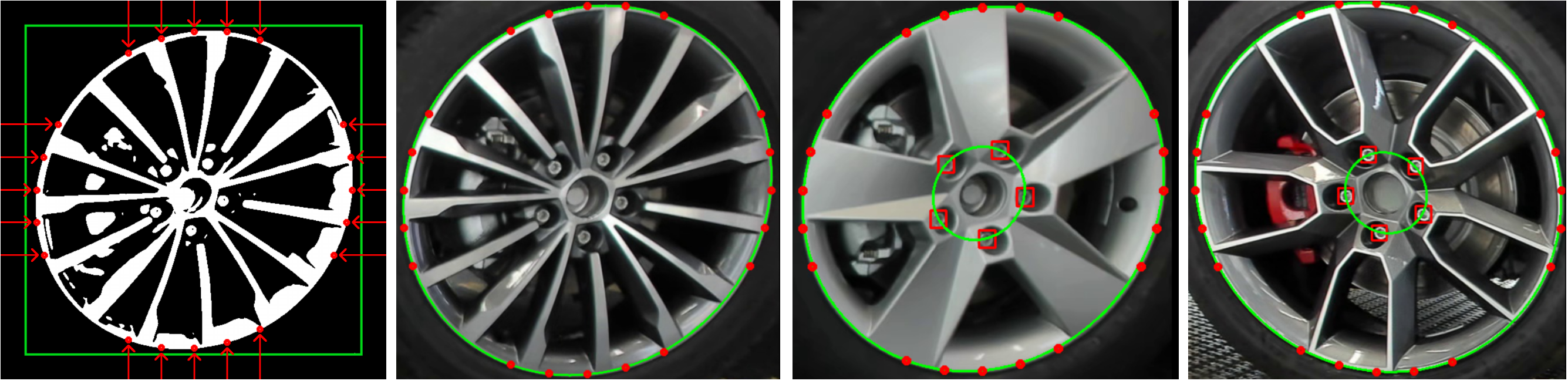}
\vspace{-0.3cm}
\caption{Detection of rim and bolt ellipses.}
\label{fig:elliptical}
\end{figure}

\vspace{-0.4cm}
\subsection{Rim size estimation}
\label{subsec:detection2}
\vspace{-0.2cm}
Since real rim dimensions are not known, we can either compare only relative sizes within a single car or use objects of known size to estimate the real rim diameter. All cars in our scenario have a pitch circle diameter of 112~mm, so the object of known size, which we have to detect, is the circle on which five bolts lie.
The cameras were not calibrated precisely and had slight discrepancies in tilt and mounting positions. The car position on the conveyor belt is not fixed, and wheels may not be perfectly perpendicular to the camera. 
Therefore, the rim circumference may not be a perfect circle but an ellipse, as shown in Figure~\ref{fig:elliptical}. 
The second detection network was trained on the RB600 set to detect the rim and bolts. The same YOLOv5s architecture as described in Section~\ref{subsec:detection1}b was used. 
The input to the second YOLO network was the bounding box of the wheel detected by the first YOLO network.
The performance of the bolt and rim detection, including the overall performance, is summarized in Table~\ref{table:detection}. Then we calculate the ellipse \cite{1998halir} on which the centers of the bolts lie.

To extract the outline of the rim, we applied Otsu's method~\cite{1979otsu} for thresholding. To find stable sample points, we cast rays from the center of each edge, spaced at 10\% of the edge length. When a ray hits a white pixel, that location is added as a sample point. We use these points to fit the second ellipse that matches the contour of the rim. Visualization of the entire procedure using rays is presented in Figure~\ref{fig:elliptical}. From these two ellipses we are able to estimate the real size of the rim.

\vspace{-0.4cm}
\subsection{Tracking}
\label{subsec:tracking}
\vspace{-0.2cm}
The tracking algorithm for car parts is of paramount significance as it improves accuracy and enhances detection speed. We use only Camera A and apply the same tracking information to Camera B due to the minor differences in horizontal coordinates. The IoU of bounding boxes conducts the data association for tracking in consecutive frames. The wheels are tracked similarly to the car and then assigned to the currently tracked car. Using the tracking information, calculating the median class for a particular rim achieves 100\% accuracy on tested videos of the total length of 10 hours containing 500 cars.

The prototype design follows the order described in Section~\ref{sec:method}. A visual representation of the primary modules and their inter-module data flow can be seen in Figure~\ref{fig:flowchart}. The procedure begins by acquiring frames from both cameras. Next, vehicle and wheel detection is performed (\ref{subsec:detection1}b). Subsequently, the rim class is predicted (\ref{subsec:classification}b), and the diameter of the rim is estimated from the bounding box surrounding the wheel (\ref{subsec:detection2}). The final step involves comparing all four wheels. The average processing time for a single input consisting of two frames, one from each camera, is 0.4 seconds for the entire pipeline. All experiments were performed on GeForce RTX 2070.

\vspace{-0.2cm}
\section{Conclusion}
\vspace{-0.2cm}
\label{sec:conclusion}
We proposed a real-time, fully automated system for rim size inspection of cars moving on the assembly line. The system consists of three steps: car and wheel detection, rim classification, and estimation of real rim dimensions. Traditional computer vision methods, such as Hough Transform and SVM with HOG features, were compared with deep learning techniques. When deep learning techniques are selected, the success rate in each intermediate step is approximately 99 percent. 
For the purpose of learning and testing, three datasets were prepared, which are publicly available for scientific purposes on Kaggle: 
\newline\href{https://www.kaggle.com/datasets/adamnovozmsk/cawdec}{https://www.kaggle.com/datasets/adamnovozmsk/cawdec}

\vfill\pagebreak
\bibliographystyle{IEEEbib}
\bibliography{strings,refs}

\end{document}